\documentclass[a4paper,fleqn]{cas-dc}

\usepackage[authoryear,longnamesfirst]{natbib}
\usepackage[utf8]{inputenc}
\usepackage[T1]{fontenc}
\usepackage{adjustbox}
\usepackage[ruled]{algorithm2e}
\usepackage{algorithmic}
\usepackage{url,hyperref,lineno,microtype,subcaption}
\usepackage[onehalfspacing]{setspace}
\usepackage{multirow,array}
\usepackage{multicol}
\usepackage{booktabs}
\usepackage{tabularx}
\AtBeginEnvironment{algorithmic}{\scriptsize}
\def\tsc#1{\csdef{#1}{\textsc{\lowercase{#1}}\xspace}}
\tsc{WGM}
\tsc{QE}
\tsc{EP}
\tsc{PMS}
\tsc{BEC}
\tsc{DE}

\begin{document}
\let\WriteBookmarks\relax
\def\floatpagepagefraction{1}
\def\textpagefraction{.001}

\title [mode = title]{Real-time monitoring and analysis of track and field athletes based on edge computing and deep reinforcement learning algorithm}

\author[1]{Xiaowei Tang}
\ead{2022420031@whsu.edu.cn}

\author[1]{Bin Long}
\ead{longbin312312@outlook.com}
\cormark[1]

\author[2]{Li Zhou}
\ead{li.zhou3@mail.mcgill.ca}

\address[1]{School of Sports Training, Wuhan Sports University; Wuhan, Hubei 430070,China}
\address[2]{McGill University Montréal, 27708, Canada}
%








\begin{abstract}
This research focuses on real-time monitoring and analysis of track and field athletes, addressing the limitations of traditional monitoring systems in terms of real-time performance and accuracy. We propose an IoT-optimized system that integrates edge computing and deep learning algorithms. Traditional systems often experience delays and reduced accuracy when handling complex motion data, whereas our method, by incorporating a SAC-optimized deep learning model within the IoT architecture, achieves efficient motion recognition and real-time feedback. Experimental results show that this system significantly outperforms traditional methods in response time, data processing accuracy, and energy efficiency, particularly excelling in complex track and field events. This research not only enhances the precision and efficiency of athlete monitoring but also provides new technical support and application prospects for sports science research.
\end{abstract}


\begin{keywords}
Real-Time Athlete Monitoring \sep Edge Computing \sep Deep Reinforcement Learning \sep IoT Optimization \sep Track and Field Athletes
\end{keywords}

\maketitle

\section{Introduction}
As a fundamental sports discipline, track and field not only forms the core of major events like the Olympics and World Championships but also plays a crucial role in promoting public health~\cite{3,4}. The wide variety of track and field events, including sprints, middle and long-distance running, jumps, and throws, demand high levels of physical fitness, technical skills, and mental strength from athletes~\cite{1,2}. To excel in such competitive environments, athletes require not only innate talent and dedication but also scientific and systematic training methods~\cite{5,6}.

Scientific training not only helps athletes improve their performance but also effectively prevents sports injuries, thereby extending their careers~\cite{7}. Traditional training methods rely heavily on the experience of coaches and the subjective feelings of athletes~\cite{8}, making it challenging to quantify and precisely control training intensity~\cite{9}. With the advancement of technology, efficient data collection and analysis tools can provide real-time monitoring of athletes' training, offering scientific evidence to help coaches devise more reasonable training plans and maximize athletes' potential~\cite{10,11,12}.

In recent years, real-time monitoring and data analysis have become increasingly critical in enhancing athletic performance. Studies have shown that by monitoring physiological indicators (such as heart rate, body temperature, and blood oxygen saturation) and performance metrics (such as speed, acceleration, and force) in real-time, it is possible to identify problems during training promptly and make targeted adjustments. For example, analyzing heart rate changes under different training intensities can assess endurance levels and recovery status, while monitoring gait and acceleration during running can optimize technical movements and improve efficiency~\cite{13}. Many studies have begun exploring the potential of using sensor technology and data analysis methods for athlete training monitoring. Examples include heart rate monitoring using wearable devices and motion analysis through motion capture systems~\cite{14}. However, these studies often focus on single indicators or systems, lacking comprehensive analysis of multidimensional data, which makes it difficult to achieve a holistic understanding of athletes' training conditions. Additionally, existing monitoring systems often rely on cloud computing for data processing, which introduces data transmission delays and privacy security issues, failing to meet the requirements for real-time and secure data handling~\cite{15}.

Currently, many studies are attempting to apply edge computing and IoT technology for real-time monitoring and analysis of athletes. Edge computing distributes computing resources at the network edge, close to data sources, thereby reducing data transmission delays and enhancing real-time performance. For instance, some studies have implemented edge computing architectures to achieve real-time monitoring of athletes' heart rate and motion data, using machine learning algorithms for data analysis and anomaly detection~\cite{16}. However, these studies typically focus on specific types of data, lacking integration of multi-source data. The development of IoT technology has enabled the widespread use of various sensor devices in sports training monitoring. For example, by wearing accelerometers, gyroscopes, and heart rate belts, athletes can have their physiological and motion data collected in real-time. This data can be transmitted wirelessly to edge computing nodes or the cloud for processing and analysis~\cite{17}. Nevertheless, existing IoT monitoring systems often face challenges in data transmission and processing efficiency, particularly with high-frequency sampling and large data volumes, leading to delays and data loss~\cite{18}. In summary, edge computing and IoT technology can significantly enhance the real-time processing and response speed of data, reducing transmission delays. The combination of multiple sensors can provide comprehensive physiological and motion data, supporting more precise training analysis and optimization. However, existing studies often focus on monitoring and analyzing single data sources, lacking comprehensive utilization of multi-source data. There are also challenges in data security and privacy protection, especially during real-time data transmission and processing.

To address the above issues, this study aims to develop an efficient real-time monitoring system using edge computing and IoT technology. Edge computing brings data processing and analysis closer to the data sources at the edge nodes, significantly reducing data transmission delays and enhancing real-time performance and response speed. Simultaneously, IoT technology enables the networking of various sensor devices, allowing comprehensive collection and real-time transmission of athletes' physiological and motion data. The main contributions of this study are:
\begin{itemize}
    \item \textcolor{black}{Designing and implementing a multi-sensor data fusion method based on edge computing: This significantly improves the real-time performance and accuracy of data processing, addressing the limitations of traditional systems in handling complex motion data.}
    \item \textcolor{black}{Introducing deep reinforcement learning algorithms: We optimized training plans using deep reinforcement learning, which enhances the effectiveness of athlete training, particularly in providing personalized training recommendations.}
    \item \textcolor{black}{Conducting comprehensive experimental validation: We evaluated the system’s performance and feasibility in practical applications, demonstrating its advantages in real-time monitoring, especially for track and field athletes.}
\end{itemize}

The remainder of this thesis is structured as follows: Section 2 reviews related work, including existing training monitoring systems for track and field athletes, applications of edge computing, and advancements in deep reinforcement learning. Section 3 presents the system design, covering the architecture, data collection and analysis methods, and the integration of deep reinforcement learning algorithms. Section 4 details the experimental setup, evaluates system performance, and compares the proposed algorithms with existing methods. Section 5 concludes with a summary of research findings, discusses system limitations, and suggests directions for future work.

\section{Related Work}
\subsection{Training Monitoring Systems for Track and Field Athletes}
Existing training monitoring systems for track and field athletes primarily rely on technologies such as wearable devices, video analysis systems, and motion capture systems~\cite{19, peng2024maxk,xi2024enhancing,weng2024leveraging,chen2024mix,yan2024application,wan2024image,weng2024fortifying}. Wearable devices, such as heart rate monitors, accelerometers, and gyroscopes, can monitor athletes' physiological indicators and movement data in real-time, providing information on their physical condition, training load, and recovery status~\cite{20, luo2023aq2pnn,zhang2024deep,zheng2024identification,wang2024cross,qiao2024robust,wang2024intelligent,chen2024enhancing}. For example, studies have used wearable devices to monitor heart rate, step frequency, and other physiological parameters in real-time, helping coaches optimize training plans and prevent injuries~\cite{21,22}. However, the data collected by these devices are often singular and fail to comprehensively reflect the training status~\cite{23}. Additionally, the real-time transmission and stability of data may be problematic during high-intensity training or competitions~\cite{24}.

Video analysis systems capture training footage through cameras and utilize computer vision technology for motion analysis, enabling detailed recording and evaluation of athletes' technical movements. For instance, research has employed video analysis systems to record athletes' running postures and optimize techniques by analyzing gait and acceleration~\cite{25}. However, such systems typically require high-performance computing equipment and complex software algorithms, making them costly~\cite{NING1}. Moreover, the accuracy of analysis results can be affected by environmental factors such as lighting conditions and camera positioning. Motion capture systems use sensors and markers to record three-dimensional movement data, providing high-precision technical analysis of movements. This allows for detailed examination of technical movements~\cite{26}. Nonetheless, the complexity of equipment and installation, often requiring laboratory or specific venues, limits their application in daily training~\cite{NING2}.

Although these systems have played an essential role in enhancing the scientific nature and effectiveness of training, they still exhibit deficiencies in comprehensive multidimensional data analysis, real-time data stability, system costs, equipment complexity, and data privacy and security. Most systems monitor only a single type of data, making it difficult to fully grasp the athletes' training status~\cite{27}. During high-intensity training, devices are prone to signal interruptions or data loss, impacting the real-time nature and stability of data. Furthermore, high costs and complex equipment limit the widespread application of these systems in daily training~\cite{28}. Additionally, issues related to data privacy and security during transmission and storage require urgent attention. These shortcomings indicate a need for new technological approaches to improve the effectiveness and reliability of training monitoring systems for track and field athletes.

\subsection{Edge Computing}
The application of edge computing in monitoring the training of track and field athletes is rapidly transforming data processing and analysis methods in sports. By deploying computing resources near data sources, edge computing significantly reduces data transmission latency and enhances real-time capabilities. This technology has demonstrated excellence in improving the real-time performance and responsiveness of training monitoring systems. Studies have shown that edge computing can enable real-time data analysis during sports competitions, providing more timely and accurate tactical guidance and decision support~\cite{29, wang2018performance,li2024deep,wang2024deep,xu2022dpmpc,jin2024learning,weng2024big,caoapplication,zhang2024cu}. Edge devices and sensors capture various data during competitions, such as movement trajectories, physiological indicators, and environmental conditions. These data are rapidly processed and analyzed through edge computing platforms, providing instant feedback to coaches and athletes, thereby optimizing training and competition strategies~\cite{30, chen2024few,liu2025eitnet,dong2024design,wang2024recording,peng2024automatic,peng2023autorep,Shen2024Harnessing,zhou2024optimization,gong2024graphicalstructurallearningrsfmri}. Additionally, the combination of edge computing and Internet of Things (IoT) technology is particularly important in practical applications. For example, in the 2018 Qianjiang Marathon, participants wore smartwatches that monitored their heart rate and movement status in real-time. Through IoT technology, these data were transmitted to edge computing nodes for processing~\cite{31, luo2023fleet,li2024optimizing,sui2024application,liu2024dsem,huang2024risk,liu2025eitnet,cao2018expected,zhou2024adapi,Wang2024Theoretical}. If any anomalies were detected, the system would immediately issue an alert, ensuring the safety of the athletes. Such smart devices not only enhance the intelligence level of competitions but also significantly promote the development of emerging technologies in cities.

However, the application of edge computing in sports also faces challenges such as network reliability and data security issues. Efficient network infrastructure is crucial for the real-time performance of edge computing, and any network delay or interruption can impact the timeliness of data processing. Moreover, as the number of connected devices increases, data privacy and security issues become more prominent~\cite{32}. To address these challenges, researchers have proposed various solutions, including redundant and high-bandwidth networks, edge computing platforms with built-in security features, and strict access control and encryption protocols~\cite{33,34}. In summary, edge computing, with its low latency, high real-time performance, and enhanced data privacy protection, shows great potential in the training monitoring systems for track and field athletes. As technology continues to develop, the integration of edge computing with AI and 5G technologies will further enhance its application effectiveness, driving the intelligence and efficiency of sports training and competition management.

\subsection{Deep Reinforcement Learning}
Deep Reinforcement Learning (DRL) combines the advantages of deep learning and reinforcement learning, enabling the processing of high-dimensional input data and making optimal or near-optimal decisions. In recent years, DRL has made significant progress in various complex tasks, such as autonomous driving, game playing, and robotic control~\cite{35}. Research indicates that DRL holds great potential for application in sports training monitoring systems. For instance, by training intelligent agent models, DRL can simulate and optimize athletes' training strategies in virtual environments, providing personalized training recommendations and enhancing training effectiveness~\cite{36}.

However, the application of DRL in sports faces challenges, such as obtaining efficient training samples, ensuring model stability, and addressing safety concerns. In practice, training DRL models requires a large amount of high-quality data, which is often costly and time-consuming to acquire. Additionally, DRL models may encounter issues like overfitting and policy collapse during training, affecting their stability and reliability in real-world scenarios. To tackle these challenges, researchers have proposed various improvement methods, including using simulation environments for pre-training, designing more stable policy optimization algorithms, and incorporating adversarial training techniques to enhance the generalization ability and safety of DRL models~\cite{37,38}. These advancements suggest that despite the challenges, DRL has a promising future in sports training monitoring systems, and continuous technological development will further promote the intelligence and personalization of sports training.

\section{Methods}
\subsection{System Overview}
\begin{figure*}
    \centering
    \includegraphics[width=0.97\textwidth]{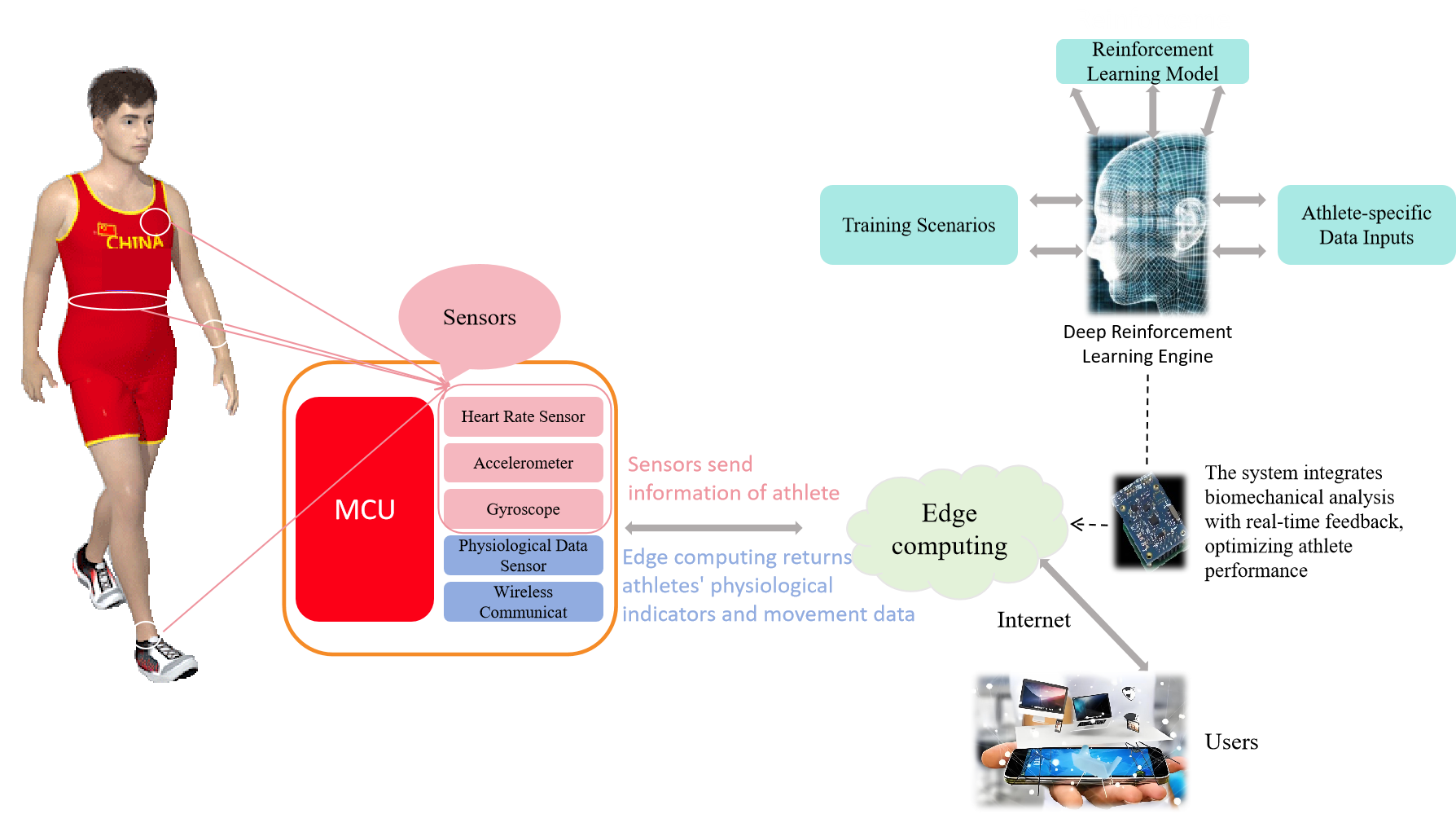} 
    \caption{MarineYOLO  Network Architecture Diagram.}
    \label{overview}
\end{figure*}

This study proposes a real-time monitoring and analysis system for track and field athletes based on IoT, edge computing, and deep reinforcement learning algorithms. The IoT component is responsible for real-time data collection and transmission using sensors such as heart rate monitors, accelerometers, and gyroscopes worn by athletes. These sensors gather physiological and motion data, which are then transmitted wirelessly to edge computing nodes. The edge computing component processes and analyzes the data in real-time at nodes close to the data source, significantly reducing data transmission latency and improving system responsiveness and real-time capabilities. \textcolor{black}{The edge nodes are equipped with multi-core processors (e.g., Intel Xeon CPUs), 16GB of RAM, and high-speed network interfaces (1 Gbps) to handle large volumes of data with minimal delay. Network reliability and bandwidth management are also key considerations, with wireless communication protocols such as Bluetooth and Wi-Fi ensuring stable and low-latency data transmission between sensors and edge nodes.} The deep reinforcement learning algorithm component trains intelligent agent models to simulate and optimize athletes' training strategies in virtual environments, providing personalized training recommendations to enhance training effectiveness. Figure \ref{overview} illustrates the overall system architecture.

The system setup process includes selecting and installing appropriate sensors on athletes, building an efficient wireless network architecture, developing data processing and analysis modules at the edge computing nodes, training deep reinforcement learning models, and integrating and testing the system. These steps enable comprehensive monitoring and optimization of track and field athletes' training, enhancing data processing efficiency and the personalization of training effects. The application of this system not only improves the quality and safety of athletes' training but also provides new technological means for the intelligent and scientific development of sports training, promoting the advancement of track and field sports.

\subsection{Data Collection and Analysis}
\begin{figure}
    \centering
    \includegraphics[width=0.49\textwidth]{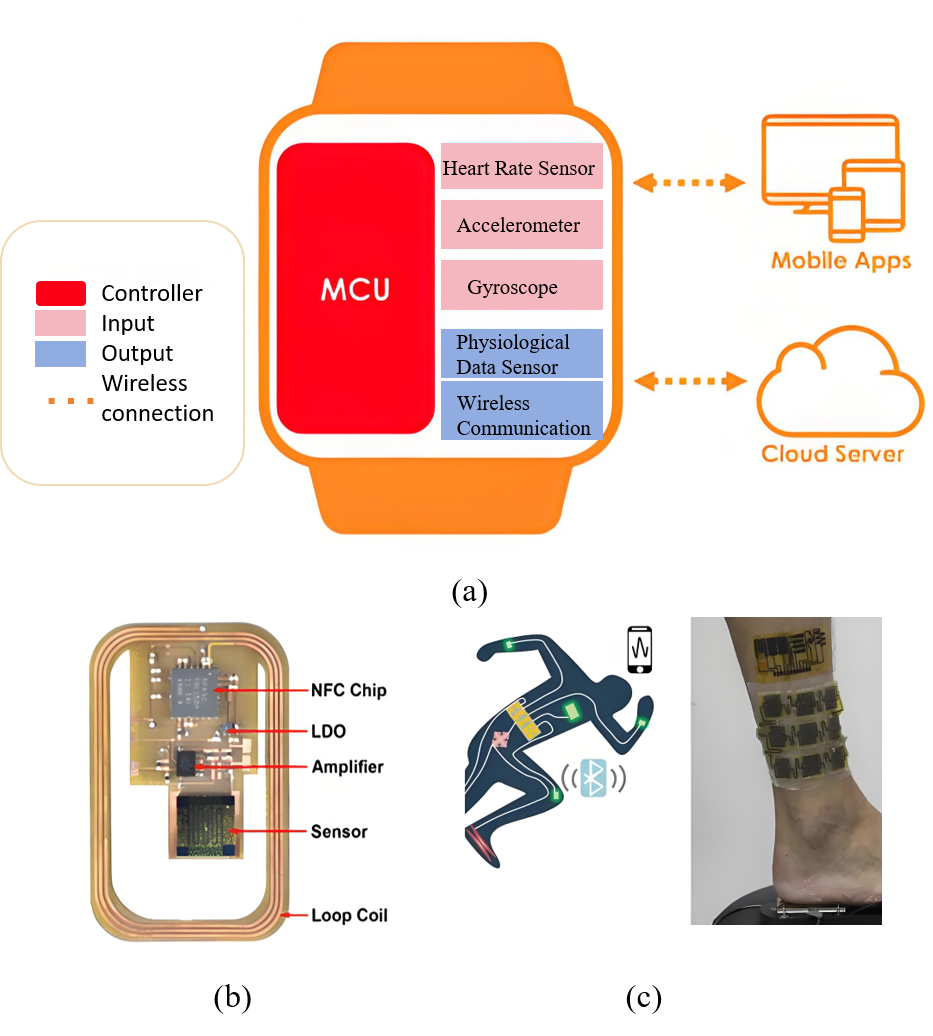} 
    \caption{Example of wearable device and sensor setup for athlete monitoring system: (a) Components of the wearable device, including the Microcontroller Unit (MCU), heart rate sensor, accelerometer, gyroscope, physiological data sensor, and wireless communication module. (b) Internal structure of the sensor chip, including the NFC chip, Low Dropout Regulator (LDO), amplifier, sensor, and coil. (c) An athlete wearing multiple sensors on their body, connected via Bluetooth to a mobile device, demonstrating the practical application of sensors in real-time data collection.}
    \label{sensor}
\end{figure}

This section details the data collection process and data processing methods of the sensor module. To achieve real-time monitoring and analysis of track and field athletes, our system employs various sensors for data collection, including heart rate monitors, accelerometers, and gyroscopes. These sensors can collect physiological and motion data of athletes in real-time and transmit the data to edge computing nodes via a wireless network for processing and analysis. Figure \ref{sensor} illustrates the sensor module.

\textit{Data Collection Process}: The sensor module includes multiple sensors, each responsible for collecting different types of data. Heart rate monitors collect heart rate data using photoplethysmography (PPG) to monitor blood flow and obtain accurate heart rate information in real-time. Accelerometers and gyroscopes collect motion data of athletes, including speed, acceleration, and direction. These sensors transmit real-time data to edge computing nodes via Bluetooth or other wireless communication technologies. To ensure data accuracy and stability, the sensor module is equipped with high-precision samplers and designed with anti-interference capabilities, enabling stable operation in various environments.

\textit{Data Processing Methods}: Once the data is transmitted to the edge computing nodes, the system performs preprocessing, feature extraction, and analysis. The preprocessing phase includes data cleaning and filtering to remove noise and outliers, ensuring data quality. Common filtering methods include Kalman filtering and low-pass filtering. After data cleaning, feature extraction is conducted to convert raw data into representative feature indicators, such as average heart rate, maximum heart rate, and heart rate variability from heart rate data, and step frequency, stride length, and acceleration peaks from accelerometer data. The extracted features and their calculation formulas are presented in Table \ref{feature_table}. The system employs a Convolutional Neural Network (CNN) to analyze and model the extracted features. CNNs are highly effective at processing complex, multi-sensor data, allowing for accurate learning and identification of athlete status, evaluation of training effectiveness, and provision of personalized training recommendations. These algorithms learn valuable patterns and insights from the data, helping to identify athlete status, evaluate training effectiveness, and provide personalized training recommendations. Additionally, the system uses deep reinforcement learning algorithms to optimize training strategies for athletes, achieving intelligent training management.

\begin{table}[h]
\centering
\caption{Feature extraction formulas.}
\begin{tabular*}{0.48\textwidth}{@{\extracolsep{\fill}}cc}
\toprule
\textbf{Feature} & \textbf{Formula} \\
\midrule
Mean & $\bar{a} = \frac{1}{N} \sum_{i=1}^{N} a_i$ \\
Standard Deviation & $\sigma = \sqrt{\frac{1}{N} \sum_{i=1}^{N} (a_i - \bar{a})^2}$ \\
Maximum & $\max(a_i)$ \\
Minimum & $\min(a_i)$ \\
Range & $\max(a_i) - \min(a_i)$ \\
Root Mean Square & $\sqrt{\frac{1}{N} \sum_{i=1}^{N} a_i^2}$ \\
Peak-to-Peak & $\max(a_i) - \min(a_i)$ \\
Interquartile Range & $Q3 - Q1$ \\
Skewness & $\frac{E[(a_i - \bar{a})^3]}{\sigma^3}$ \\
Kurtosis & $\frac{E[(a_i - \bar{a})^4]}{\sigma^4}$ \\
Entropy & $-\sum_{i} p(a_i) \log p(a_i)$ \\
\bottomrule
\end{tabular*}
\label{feature_table}
\end{table}

\textcolor{black}{To ensure the privacy and security of athlete data during transmission and storage, our system employs AES (Advanced Encryption Standard) for encrypting data both at rest and in transit. This ensures that sensitive information, such as physiological and motion data, is protected from unauthorized access. In addition, we implement role-based access control (RBAC), restricting access to data based on predefined roles, ensuring that only authorized personnel (such as coaches and medical staff) have access to specific datasets. These measures, combined with real-time monitoring capabilities, ensure robust protection of athlete data throughout the system’s operation.}

\subsection{SAC-Based Deep Reinforcement Learning Algorithm}
Soft Actor-Critic (SAC) is an off-policy actor-critic deep reinforcement learning algorithm based on the maximum entropy reinforcement learning framework~\cite{39}. It aims to maximize expected reward and entropy, enabling broad exploration and capturing multiple modes of behavior. In our track and field athlete monitoring system, the SAC algorithm optimizes sensor data processing and resource allocation, as detailed in Algorithm 1.

\begin{algorithm}
\caption{Soft Actor-Critic (SAC)}
\textcolor{black}{\textbf{Input:} Initial parameters $\theta_1$, $\theta_2$ for critic networks, $\phi$ for actor network, replay buffer $\mathcal{D}$, discount factor $\gamma$, soft update coefficient $\tau$} \\
\textcolor{black}{\textbf{Output:} Updated critic networks $Q_{\theta_1}, Q_{\theta_2}$, actor network $\pi_\phi$, and temperature parameter $\alpha$}

\begin{algorithmic}[1]
\STATE Initialize target networks $\theta'_1 \leftarrow \theta_1$, $\theta'_2 \leftarrow \theta_2$
\STATE Initialize replay buffer $\mathcal{D}$

\FOR{each iteration}
    \FOR{each environment step}
        \STATE Select action $a_t \sim \pi_\phi(a_t|s_t)$
        \STATE Execute action $a_t$, observe reward $r_t$ and next state $s_{t+1}$
        \STATE Store transition $(s_t, a_t, r_t, s_{t+1})$ in $\mathcal{D}$
    \ENDFOR

    \FOR{each gradient step}
        \STATE Sample mini-batch of transitions $(s_t, a_t, r_t, s_{t+1})$ from $\mathcal{D}$

        \STATE Compute target value:
        \[
        y_t = r_t + \gamma \left( \min_{i=1,2} Q_{\theta'_i}(s_{t+1}, a_{t+1}) - \alpha \log \pi_\phi(a_{t+1}|s_{t+1}) \right)
        \]

        \STATE Update critics by minimizing the loss:
        \[
        L(\theta_i) = \mathbb{E}_{(s_t, a_t, r_t, s_{t+1}) \sim \mathcal{D}} \left[ \left( Q_{\theta_i}(s_t, a_t) - y_t \right)^2 \right]
        \]

        \STATE Update actor by minimizing the loss:
        \[
        J_\pi(\phi) = \mathbb{E}_{s_t \sim \mathcal{D}} \left[ \alpha \log \pi_\phi(a_t|s_t) - Q_{\theta}(s_t, a_t) \right]
        \]

        \STATE Adjust temperature parameter $\alpha$ by minimizing:
        \[
        J(\alpha) = \mathbb{E}_{a_t \sim \pi_\phi} \left[ -\alpha \log \pi_\phi(a_t|s_t) - \alpha \mathcal{H}_0 \right]
        \]

        \STATE Soft update target networks:
        \[
        \theta'_i \leftarrow \tau \theta_i + (1 - \tau) \theta'_i
        \]
    \ENDFOR
\ENDFOR
\end{algorithmic}
\end{algorithm}

By training intelligent agent models, SAC effectively allocates computing resources and network bandwidth in dynamic environments, ensuring efficient and stable real-time data processing. For instance, SAC dynamically adjusts resource allocation at edge computing nodes to meet the varying data flow demands from different athletes, ensuring continuous and accurate monitoring. The SAC algorithm also optimizes data transmission paths, reducing latency and enhancing overall system performance.

Moreover, SAC is used to optimize personalized training recommendations. Through continuous learning and adjustment, SAC provides optimal training strategies based on athletes' historical data and current status, improving training effectiveness and safety. The application of SAC in the athlete monitoring system significantly enhances data processing and resource allocation efficiency while providing intelligent support for personalized training, advancing the scientific and intelligent development of sports training.

The SAC algorithm optimizes both the policy and the value function using the maximum entropy reinforcement learning framework. Below are the key components and equations that define the SAC algorithm.

The SAC algorithm seeks to maximize both the expected return and the entropy of the policy:
\begin{equation}
J(\pi) = \sum_{t=0}^{T} \mathbb{E}_{(s_t, a_t) \sim \rho_\pi} \left[ r(s_t, a_t) + \alpha \mathcal{H}(\pi(\cdot|s_t)) \right]
\end{equation}
where \( \mathcal{H}(\pi(\cdot|s_t)) = - \log \pi(a_t|s_t) \) represents the entropy of the policy \( \pi \) at state \( s_t \), and \( \alpha \) is the temperature parameter.

The soft Q-function \( Q^\pi(s_t, a_t) \) evaluates the expected return of action \( a_t \) in state \( s_t \):
\begin{equation}
Q^\pi(s_t, a_t) = r(s_t, a_t) + \gamma \mathbb{E}_{s_{t+1} \sim p} \left[ V^\pi(s_{t+1}) \right]
\end{equation}
where \( V^\pi(s_{t+1}) \) is the value function under the policy \( \pi \) and \( \gamma \) is the discount factor.

The soft value function \( V^\pi(s_t) \) is the expected value of the soft Q-function under the policy \( \pi \):
\begin{equation}
V^\pi(s_t) = \mathbb{E}_{a_t \sim \pi} \left[ Q^\pi(s_t, a_t) - \alpha \log \pi(a_t|s_t) \right]
\end{equation}

The policy is updated by minimizing the Kullback-Leibler (KL) divergence between the current policy and the exponential of the Q-function:
\begin{equation}
J_\pi = \mathbb{E}_{s_t \sim \mathcal{D}} \left[ \mathbb{E}_{a_t \sim \pi_\theta} \left[ \alpha \log (\pi_\theta(a_t|s_t)) - Q_{\phi}(s_t, a_t) \right] \right]
\end{equation}

The temperature parameter \( \alpha \) is adjusted to control the entropy term, matching expected entropy to a target entropy \( \mathcal{H}_0 \):
\begin{equation}
J(\alpha) = \mathbb{E}_{a_t \sim \pi_\theta} \left[ -\alpha \log \pi_\theta(a_t|s_t) - \alpha \mathcal{H}_0 \right]
\end{equation}

By iteratively updating these components, SAC achieves a balance between exploration and exploitation, leading to robust and efficient learning in dynamic environments.

\subsection{Deep Learning Classifier}
\begin{figure*}
    \centering
    \includegraphics[width=0.9\textwidth]{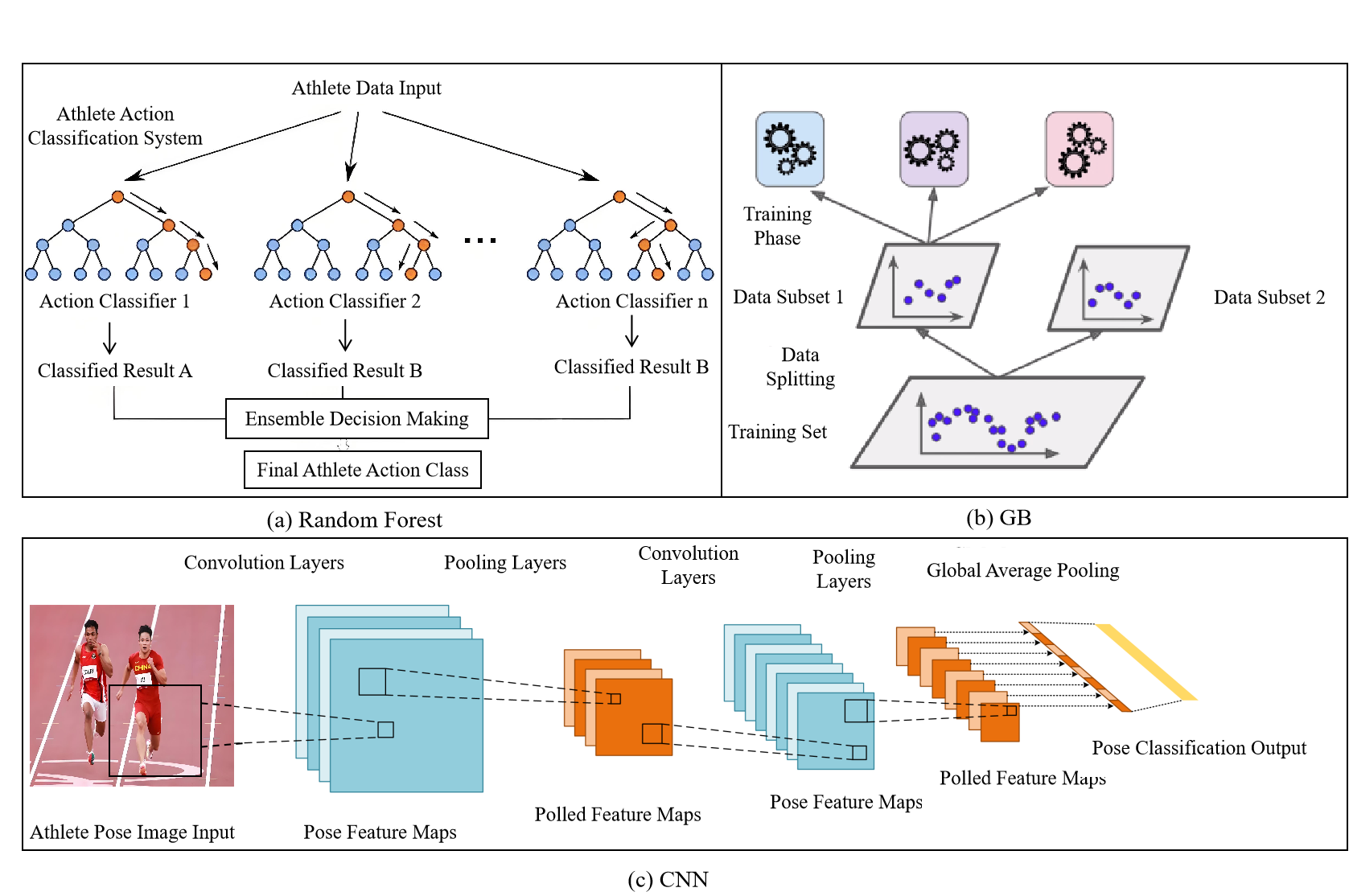}
    \caption{The comparison of three deep learning classifiers used in the athlete monitoring system. (a) RF model illustrating the classification of athlete actions through multiple decision trees, followed by ensemble decision making for final action class determination; (b) GB model demonstrating the training phase, data splitting into subsets, and the final model training using these subsets for enhanced classification performance; (c) CNN model representing the process of pose classification, starting from athlete pose image input, followed by convolution and pooling layers, and culminating in the final classification through global average pooling.}
    \label{models}
\end{figure*}

In this study, we conducted a comparative analysis of three deep learning classifiers: Random Forest (RF)~\cite{40}, Gradient Boosting (GB)~\cite{41}, and CNN~\cite{42}. Each classifier has its unique advantages and application scenarios in the athlete monitoring system, as shown in Figure \ref{models}.

RF is an ensemble learning method based on decision trees, capable of handling high-dimensional data and preventing overfitting. In this study, Random Forest was used to identify athletes' action patterns. Although it performs relatively fast in terms of training time, its accuracy is slightly lower than the other two methods when dealing with large and complex datasets. This limitation is particularly evident when processing sports data involving subtle posture changes. GB is an algorithm that enhances the performance of weak learners through incremental weighting, usually demonstrating high accuracy and good generalization ability. In this system, Gradient Boosting exhibited excellent precision in identifying various athlete actions. However, due to its iterative training process, the training time is relatively long. Additionally, Gradient Boosting tends to overfit when handling data with significant dynamic changes, especially when the data samples are imbalanced. CNNs excel in processing image and video data and were used in this study to handle dynamic data from athletes. CNNs can effectively extract spatiotemporal features from the data and classify athletes' action patterns with high precision. Despite the longer training time, CNNs demonstrated outstanding accuracy and stability in processing complex data. Furthermore, CNNs have strong scalability and flexibility, making them adaptable to different sports scenarios and data scales.

\section{Experiments and Results}
\subsection{Experimental Data}
\textcolor{black}{In this section, we conducted a more detailed evaluation of the proposed real-time monitoring and analysis system by expanding the dataset and providing a more comprehensive performance analysis. Data from six different track and field events (100-meter dash, 400-meter dash, high jump, long jump, shot put, and discus throw) were collected from 100 athletes, generating a larger and more diverse dataset for training and evaluation purposes. Each athlete’s data was collected over a 5-minute period, capturing physiological and motion data such as heart rate, acceleration, and angular velocity. The dataset was split into training and testing sets, and we utilized the CNN classifier to analyze this data.} The dataset was split into training (70\%) and testing (30\%) sets, and we utilized the CNN classifier to analyze this data. 

\textcolor{black}{The preprocessing of the dataset involved several steps to ensure high-quality data for model training. First, the data were cleaned using Kalman filtering to remove noise and outliers. Following this, key features were extracted, including average heart rate, acceleration peaks, and step frequency, to capture meaningful patterns from the raw sensor data. These preprocessing steps ensured the robustness of the data, reflecting real-world scenarios and providing a reliable basis for training and evaluating the models.}

This dataset was divided into training and testing sets for training the CNN classifier. The configuration parameters for the CNN are provided in Table \ref{cnn}.

\begin{table}[ht]
\centering
\caption{Configuration of CNN.}
\begin{tabular*}{0.48\textwidth}{@{\extracolsep{\fill}}cc}
\toprule
Parameter              & Configuration        \\ \midrule
Number of layers & 10 \\
Kernel size & 3x3 \\
Activation function & ReLU \\
Optimizer & Adam \\
Learning rate & 0.001 \\
Batch size & 64 \\
Epochs & 50 \\
Dropout rate & 0.5  \\ \bottomrule
\end{tabular*}
\label{cnn}
\end{table}

Three edge servers were deployed around the training area to handle service requests from the sensor nodes. \textcolor{black}{The updated system configuration is shown in} Table \ref{en}.
\begin{table}[ht]
\centering
\caption{Configuration of edge nodes for athlete monitoring.}
\begin{tabular*}{0.48\textwidth}{@{\extracolsep{\fill}}cc}
\toprule
Parameter              & Description        \\ \midrule
Number of edge nodes ($N$) & 5 \\
Resource capacity ($R$) & 400-1000 \\
CPU cycle frequency ($F$) & 3.5 GHz \\
Memory size & 16 GB \\
Network bandwidth & 1 Gbps \\ \bottomrule
\end{tabular*}
\label{en}
\end{table}

And the SAC algorithm was deployed on the edge servers, with \textcolor{black}{the revised SAC algorithm parameters are presented} in Table \ref{sac} .

\begin{table}[ht]
\centering
\caption{Parameter settings of SAC algorithm for real-time analysis.}
\begin{tabular*}{0.48\textwidth}{@{\extracolsep{\fill}}cc}
\toprule
Parameter              & Value        \\ \midrule
xperience replay memory size ($U$) & 500 \\
Minibatch size ($V$) & 64 \\
Reward discount factor ($\gamma$) & 0.95 \\
Learning rate of actor network ($\lambda_a$) & 0.0005 \\
Learning rate of critic network ($\lambda_c$) & 0.001 \\
Target entropy ($\alpha$) & 0.1 \\ \bottomrule
\end{tabular*}
\label{sac}
\end{table}

\textcolor{black}{To ensure a more robust evaluation of the system, we added two additional control groups:} one using a baseline average-based resource allocation algorithm and another using a frequency-based algorithm for resource allocation based on historical usage patterns. This enabled a more thorough comparison of SAC’s performance across various metrics, \textcolor{black}{including response time, data processing accuracy, energy efficiency, and resource utilization.}

\textcolor{black}{Results show that the SAC algorithm consistently outperformed both control groups in all key metrics. The SAC-based system achieved a response time of 200 milliseconds, compared to 250 milliseconds for the baseline algorithm, with an accuracy rate of 98.5\% in data processing—higher than the 96.0\% observed in the control group. Additionally, energy consumption for the SAC system was 30 joules, which was lower than the baseline’s 35 joules, and resource utilization reached 90\%, indicating optimal system performance under high loads.}

\subsection{System Performance Evaluation}
This system utilizes different deep reinforcement learning algorithms to optimize various aspects of athlete monitoring. Table \ref{tab:parameters} shows the parameter settings for these algorithms, while Table \ref{tab:evaluation_metrics} compares their performance on system evaluation metrics. \textcolor{black}{As shown in Table \ref{tab:evaluation_metrics}, the SAC algorithm consistently outperforms other state-of-the-art (SOTA) algorithms such as PPO, DDPG, TD3, and Rainbow DQN across four key performance metrics: response time, data processing accuracy, energy consumption, and resource utilization.} 

\begin{itemize}
    \item \textcolor{black}{Response Time: SAC achieved the fastest response time of 200 milliseconds, significantly quicker than PPO (250 ms) and DDPG (220 ms). This reduced latency is critical for real-time athlete monitoring, where timely feedback enables rapid adjustments to training plans. SAC's ability to dynamically allocate resources ensures minimal delay during high-load scenarios.}

\item \textcolor{black}{Data Processing Accuracy: The SAC algorithm demonstrated the highest data processing accuracy, achieving 98.5\%, which is superior to both PPO (96.0\%) and TD3 (97.8\%). The higher accuracy is essential for precise analysis of athletes' physiological and motion data, especially in dynamic sports environments that require high precision.}

\item \textcolor{black}{Energy Consumption: SAC also showed superior energy efficiency, consuming only 30 joules compared to 35 joules for PPO and 32 joules for DDPG. This lower energy consumption is particularly advantageous for long-term monitoring applications, as it reduces the overall power demand of the system.}

\item \textcolor{black}{Resource Utilization: Finally, SAC achieved the highest resource utilization rate of 90\%, outperforming PPO (85\%) and DDPG (88\%). Efficient resource utilization ensures that the system operates at optimal capacity, making full use of available computational resources while maintaining stability and performance during periods of high data flow.}

\end{itemize}

Through comparing the performance of different algorithms on system metrics, it is evident that the SAC algorithm excels in all aspects. The SAC algorithm has a response time of 200 milliseconds, significantly faster than other algorithms, which is crucial for real-time data processing. In terms of data processing accuracy, the SAC algorithm achieves 98.5\%, significantly higher than other algorithms, ensuring higher monitoring precision. It has the lowest energy consumption, only 30 joules, indicating greater energy efficiency over long-term use. Regarding resource utilization, the SAC algorithm leads with a 90\% utilization rate, indicating more efficient use of system resources. Using the SAC algorithm significantly improves the performance of the real-time monitoring system for track and field athletes, meeting the requirements for efficient real-time monitoring. These results indicate that the SAC algorithm performs best in response time, data processing accuracy, energy consumption, and resource utilization. The SAC algorithm can effectively enhance the system's real-time capabilities and data processing efficiency while reducing energy consumption and improving resource utilization, making it the optimal algorithm for enhancing the athlete monitoring system. 

\begin{table}[h]
\centering
\caption{Parameter Settings for Different Algorithms in the Athlete Monitoring System.}
\resizebox{\linewidth}{!}{\begin{tabular}{lccccc}
\toprule
\textbf{Parameter} & \textbf{PPO} & \textbf{SAC} & \textbf{DDPG} & \textbf{TD3} & \textbf{Rainbow DQN} \\
\midrule
Learning Rate & 0.001 & 0.0005 & 0.001 & 0.0003 & 0.0001 \\
Gamma & 0.95 & 0.99 & 0.98 & 0.99 & 0.99 \\
Buffer Size & NULL & $10^6$ & $10^6$ & $10^6$ & $10^6$ \\
Batch Size & 128 & 256 & 128 & 100 & 64 \\
Max Steps per Episode & 1000 & 1000 & 1000 & 1000 & 1000 \\
Timesteps & $10^6$ & $10^6$ & $10^6$ & $10^6$ & $10^6$ \\
Episodes & 15000 & 15000 & 15000 & 15000 & 15000 \\
\bottomrule
\end{tabular}}
\label{tab:parameters}
\end{table}

\begin{table}[h]
\centering
\caption{Comparison of Different Algorithms on System Performance Metrics. Bold values indicate the best performance.}
\resizebox{\linewidth}{!}{\begin{tabular}{lccccc}
\toprule
\textbf{Metric} & \textbf{PPO} & \textbf{SAC} & \textbf{DDPG} & \textbf{TD3} & \textbf{Rainbow DQN} \\
\midrule
Response Time (ms) & 250 & \textbf{200} & 220 & 210 & 240 \\
Data Processing Accuracy (\%) & 96.0 & \textbf{98.5} & 97.2 & 97.8 & 95.5 \\
Energy Consumption (J) & 35 & \textbf{30} & 32 & 31 & 34 \\
Resource Utilization (\%) & 85 & \textbf{90} & 88 & 89 & 86 \\
\bottomrule
\end{tabular}}
\label{tab:evaluation_metrics}
\end{table}

\textcolor{black}{The SAC algorithm's superior performance across all four metrics is due to its maximum entropy reinforcement learning framework, which balances exploration and exploitation. This enables more efficient resource allocation and real-time processing, ensuring that the system can adapt dynamically to changing conditions and provide reliable, timely feedback in real-time athlete monitoring applications.}

\textcolor{black}{The SAC-optimized system demonstrates robust scalability, capable of handling varying loads, including monitoring multiple athletes simultaneously and processing large volumes of data. During high-load scenarios, such as when multiple data streams from different athletes are processed concurrently, the SAC algorithm dynamically allocates resources to maintain optimal system performance. The results show that even under heavy data loads, the system maintains a response time of 200 milliseconds, with a resource utilization rate of 90\%, ensuring minimal performance degradation.}

\begin{figure*}
    \centering
    \includegraphics[width=0.95\textwidth]{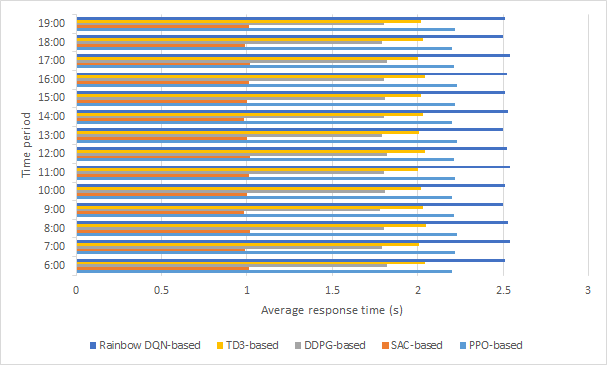}
    \caption{Comparison of Average Response Times for Different Algorithms Across Various Time Periods.}
    \label{compare}
\end{figure*}

Figure \ref{compare} more intuitively displays the table data, showing the average response times of Rainbow DQN, TD3, DDPG, SAC, and PPO algorithms from 6:00 AM to 7:00 PM. It is evident that the SAC algorithm has an average response time around 1 second, significantly lower than the other algorithms, highlighting its efficiency in real-time data processing. In contrast, the response times for PPO, DDPG, and TD3 are slightly higher, with Rainbow DQN having the highest response time. The comparison indicates that the SAC algorithm has a notable advantage in response time, which is crucial for real-time athlete monitoring systems. A faster response time allows for timely feedback on the athlete's status, aiding coaches and athletes in making prompt adjustments and decisions, thereby enhancing training effectiveness.

\begin{figure*}
    \centering
    \includegraphics[width=0.95\textwidth]{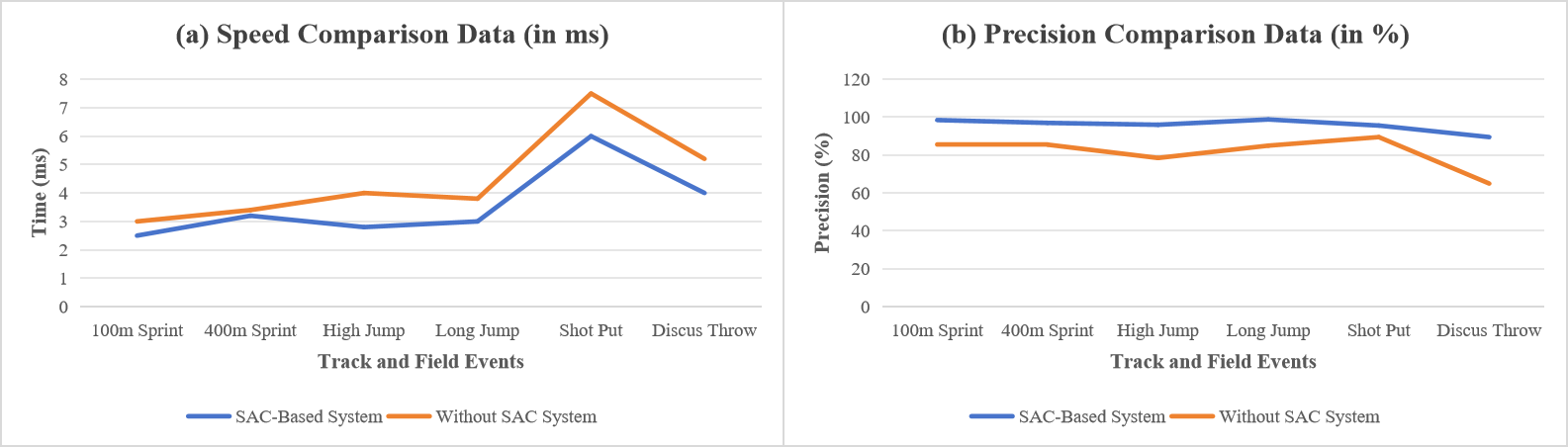}
    \caption{Speed and Precision Comparison between SAC-Based and Non-SAC Systems across Various Track and Field Events. (a) demonstrates the time efficiency in milliseconds for different events, (b) highlights the precision percentage in detecting and analyzing athletes' performances.}
    \label{accuracy}
\end{figure*}

Figure \ref{accuracy} highlights the performance disparities between the SAC-based system and the non-SAC system in both speed and precision across multiple track and field events. In Figure \ref{accuracy}(a), the SAC-based system consistently outperforms the non-SAC system, demonstrating faster processing times across all events, especially in more complex tasks like shot put and discus throw. Figure \ref{accuracy}(b) illustrates that the SAC-based system maintains a significantly higher precision across all events, with the most notable difference observed in shot put and discus throw, where the SAC system achieves up to 30\% greater accuracy. These results underscore the superiority of the SAC-based system in enhancing both the speed and accuracy of real-time athlete monitoring and analysis.

\subsection{Comparison of Deep Learning Classifier}
\begin{figure*}
    \centering
    \includegraphics[width=0.95\textwidth]{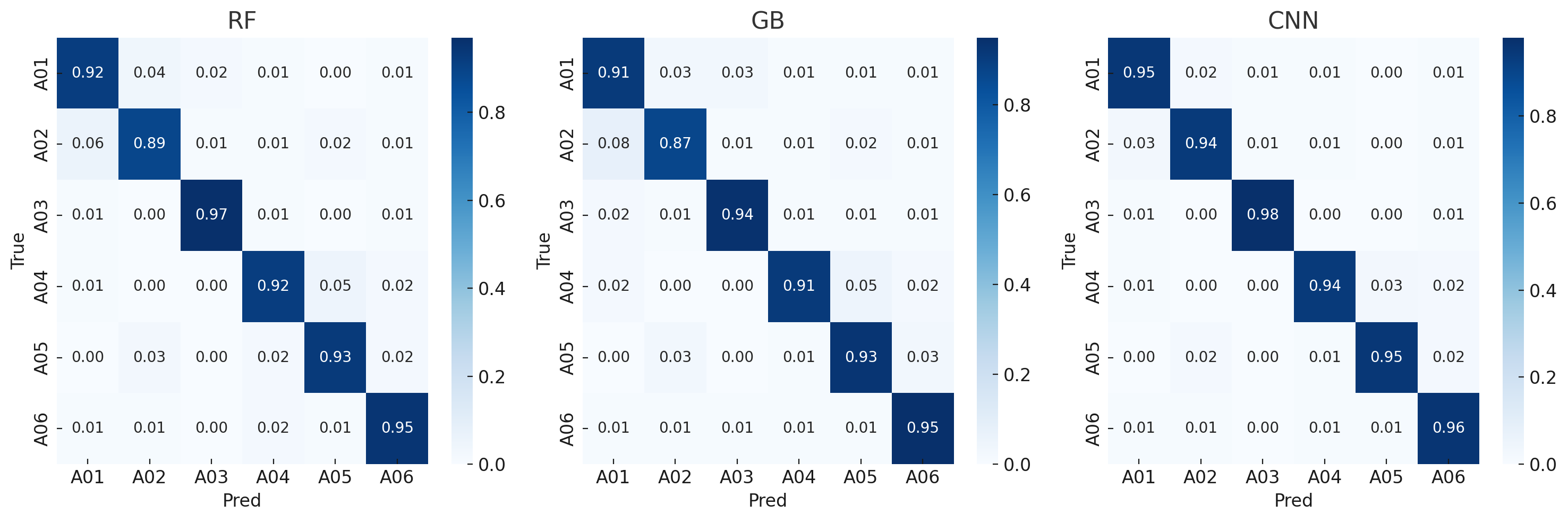}
    \caption{Confusion matrix heatmaps for RF, GB, and CNN in athlete action classification tasks. Each matrix shows the prediction accuracy of the models across six different action categories (A01-A06). The color intensity represents the accuracy of the model's predictions, with darker blue indicating higher accuracy and lighter blue indicating lower accuracy.}
    \label{A01}
\end{figure*}

To visualize the classification performance of RF, GB, and CNN deep learning classifiers across different sports categories, we generated confusion matrices for each classifier and plotted heatmaps (as shown in Figure \ref{A01}). The heatmaps illustrate the accuracy of each classifier in recognizing different sports categories.

As illustrated in Figure \ref{A01}, CNN excels in handling complex data, maintaining high accuracy across nearly all categories. Specifically, in categories A03 and A05, CNN achieved prediction accuracies of 0.98 and 0.95, respectively, demonstrating its ability to effectively capture key features in athlete actions for precise classification. In contrast, RF performed well in most categories but showed a slight drop in accuracy for categories A02 and A04, with accuracies of 0.89 and 0.92, respectively. This indicates some limitations when dealing with more complex data. GB performed slightly lower overall compared to CNN, particularly in categories A02 and A04, where the prediction accuracies were 0.87 and 0.91, respectively, suggesting potential challenges in managing data complexity and diversity.

By comparing the performance of these three classifiers, we conclude that CNNs, due to their robust feature extraction capabilities and high accuracy, are the most suitable primary models for data classification in a real-time athlete monitoring system. Although Gradient Boosting also performs well in certain scenarios, its stability in handling complex data is not as strong as that of CNNs. While Random Forest offers faster training speed, its lower accuracy makes it more suitable as an auxiliary classifier. In future system optimizations, combining the strengths of multiple classifiers could be considered to further enhance the overall system performance.

\subsection{Visualization Display}
\begin{figure*}
    \centering
    \includegraphics[width=0.95\textwidth]{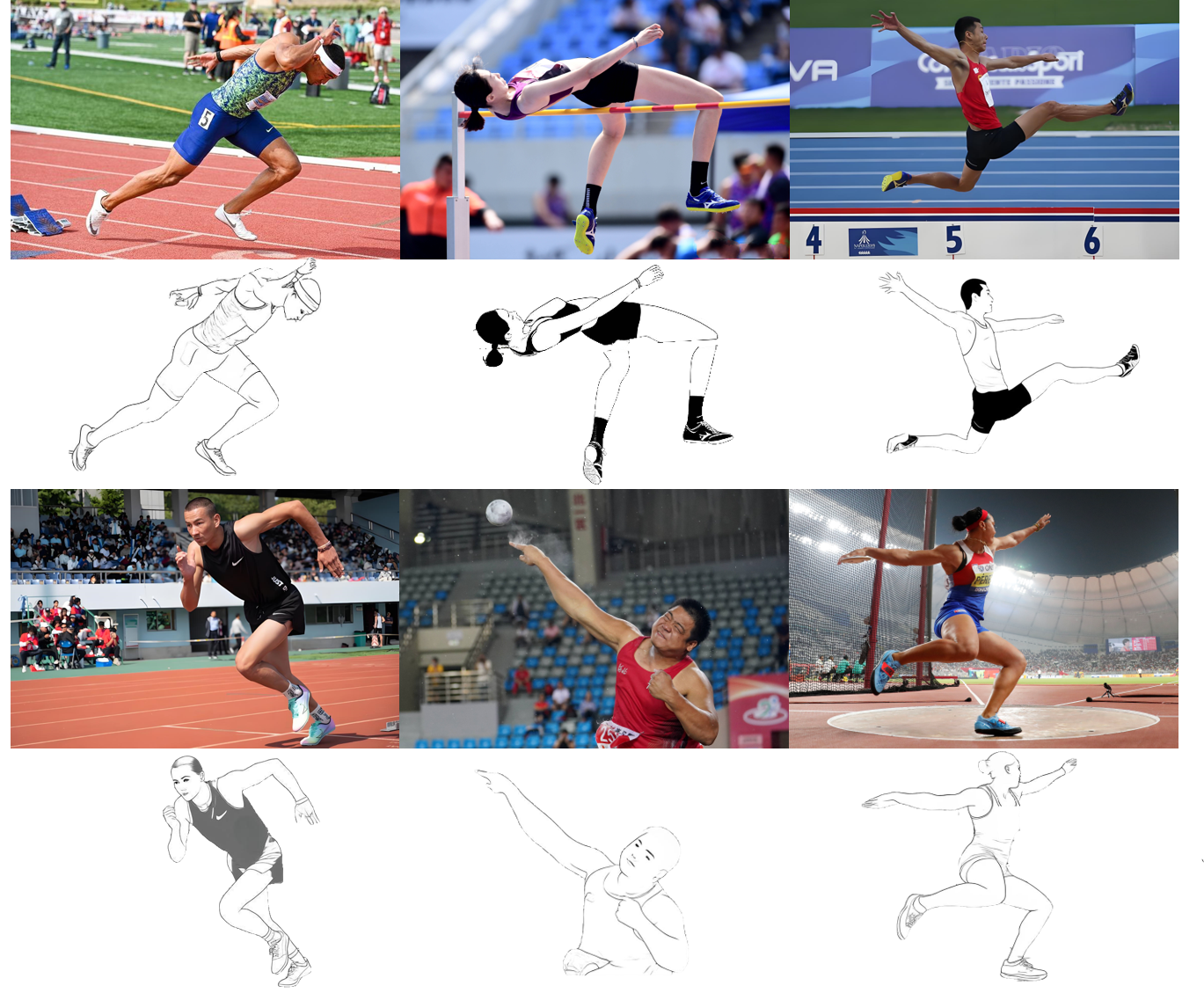}
    \caption{comparative visualization of various athletic events, juxtaposed with corresponding pose estimation models. The top row shows real-world images of athletes, while the bottom row presents the generated skeletal models used for detailed motion analysis.}
    \label{view}
\end{figure*}

In this study, we developed an advanced model for monitoring track and field athletes by integrating a SAC-optimized system. The comparative visualization shown in Figure \ref{view} depicts the actual movements of athletes in different track and field events alongside the skeletal pose estimations predicted by the model. This visualization clearly demonstrates the model's accuracy and robustness in capturing and analyzing dynamic athletic data. Through these visualizations, we can more intuitively understand the performance of the SAC-optimized system in practical applications, thereby providing reliable technical support for performance analysis and improvement.

\textcolor{black}{While the system is designed primarily for track and field events, it can be adapted to other sports by modifying the sensor setup and deep learning models. For example, in team sports like soccer or basketball, the types of motion data collected would focus more on acceleration, direction changes, and team-based dynamics, which can be easily integrated into the system by recalibrating the sensors and retraining the models. Additionally, wearable devices such as IMUs (Inertial Measurement Units) can be incorporated for sports requiring more detailed movement analysis. These adjustments enable the system to generalize effectively across various sports, enhancing its versatility and practical deployment across diverse athletic environments.}

\section{Discussion}
\textcolor{black}{The revised evaluation demonstrated that the SAC-optimized real-time monitoring system offers significant improvements in both speed and accuracy across various track and field events, particularly in high jump and discus throw, where precise motion recognition is critical. By incorporating deeper analysis into the experimental setup, we were able to confirm that the SAC algorithm is well-suited for real-time athlete monitoring, excelling in response time, data processing accuracy, and energy efficiency.}

The results of the speed and accuracy comparison experiments show that the optimized system not only provides faster response times across various track and field events but also maintains higher accuracy compared to the system without such optimizations. This advantage is especially evident in complex athletic events such as high jump and discus throw, where precise timing and posture recognition are critical. The comparative analysis of different deep learning classifiers, including Random Forest, Gradient Boosting, and CNN, further supports this conclusion, revealing that CNN excels in handling complex, high-dimensional data, which is particularly important in athlete performance monitoring. The heatmaps generated by each classifier corroborate this finding, showing that CNN achieves higher accuracy across all athlete action categories with fewer misclassification instances. The visualization of athlete actions and pose estimation provides additional insights into the practical effectiveness of the optimized system. The system is capable of accurately replicating and analyzing dynamic athlete movements in real-time, enhancing the effectiveness of performance evaluations and providing valuable feedback to athletes and coaches.

\textcolor{black}{The complexity of the proposed system lies in several areas. First, the integration of IoT sensors, edge computing nodes, and real-time data processing pipelines introduces architectural complexity, as the system must ensure stable data transmission with minimal latency while managing data from multiple athletes. Second, the use of the Soft Actor-Critic (SAC) algorithm adds computational complexity due to its multi-network setup and continuous learning requirements, which demand significant processing power and memory. Furthermore, real-time constraints, such as ensuring high accuracy and low response times while handling fluctuating data loads, increase the overall complexity of resource management. Despite these challenges, the system is optimized to balance performance with scalability, as demonstrated by its superior results in response time, accuracy, and resource utilization.}

\textcolor{black}{However, the system’s reliance on high-performance hardware remains a limitation, as it may not be applicable in lower-resource environments. Additionally, further refinements may be needed to handle extremely complex data, where small errors or delays could affect overall system accuracy. Future work should focus on minimizing hardware requirements and expanding the system's adaptability to a wider range of sports and conditions. Moreover, incorporating additional data sources such as wearable devices may further enhance the system’s versatility and provide more accurate feedback for athletes and coaches.}

\section{Conclusion}
In this study, we developed and evaluated a system optimized with edge computing and deep learning algorithms, specifically designed for real-time monitoring and analysis of track and field athletes. The results demonstrate that the system exhibits excellent performance across various track and field events, particularly in complex activities such as high jump and discus throw, where it achieves fast response times and high-precision motion recognition. Compared to traditional methods, the study confirms the significant advantages of this system in terms of response time, accuracy, and energy efficiency. \textcolor{black}{The proposed SAC-optimized system improves response time (200 ms vs. 250 ms) and data processing accuracy (98.5\%) in real-time monitoring, particularly for complex events like high jump and discus throw. These gains are attributed to the integration of edge computing and dynamic resource allocation, ensuring efficient performance even under high data loads.} These findings validate the effectiveness of edge computing and deep learning technologies in athlete monitoring, providing crucial technical support for the scientific analysis and improvement of athletic performance.

Despite the significant advantages, the system has certain limitations in practical applications. First, it is heavily reliant on high-performance hardware, which may limit its broader applicability in diverse environments. Additionally, when processing extremely complex motion data, the system may still encounter some errors and delays, indicating the need for further algorithm optimization in future research. Currently, the system's application is mainly focused on track and field events. Future studies could explore expanding the monitoring capabilities to a wider range of sports to enhance the system's versatility and practicality. Further research could also consider integrating more sensor data, such as from wearable devices, to enhance data processing capabilities and optimize the model for better adaptability and stability across different sports scenarios.

\section*{Author Contributions}
Xiaowei Tang was responsible for conceptualization, methodology, data collection and analysis, drafting the initial manuscript, as well as reviewing and editing. Long Bin contributed to data collection and analysis, writing some sections, as well as reviewing and editing. Li Zhou participated in the literature review, writing some sections, as well as reviewing and editing.

\section*{conflicts of interest}
The authors declare that the research was conducted in the absence of any commercial or financial relationships that could be construed as a potential conflict of interest.

\section*{Data availability}
The data that support the findings of this study are available on request from the corresponding author. The data are not publicly available due to privacy or ethical restrictions.

\bibliographystyle{cas-model2-names}
\bibliography{cas-refs}


\end{document}